\pdfoutput=1

\documentclass[11pt]{article}

\usepackage[final]{acl}

\usepackage{times}
\usepackage{latexsym}

\usepackage[T1]{fontenc}

\usepackage[utf8]{inputenc}

\usepackage{microtype}

\usepackage{inconsolata}

\usepackage{graphicx}

\usepackage{url}            
\usepackage{booktabs}       
\usepackage{amsfonts}       
\usepackage{nicefrac}       
\usepackage{microtype}      
\usepackage{xcolor}         
\usepackage{natbib}
\usepackage{bbm}
\usepackage{amsmath}
\usepackage[]{mdframed}
\usepackage{float}
\usepackage{hyperref}

\title{VLind-Bench: Measuring Language Priors in Large Vision-Language Models}

\author{%
Kang-il Lee$^1$ \quad Minbeom Kim$^2$ \quad Seunghyun Yoon$^3$ \quad Minsung Kim$^1$\\
\textbf{Dongryeol Lee}$^1$ \quad \textbf{Hyukhun Koh}$^2$ \quad \textbf{Kyomin Jung}$^{1,2}\thanks{Corresponding authors.}$\\
$^1$Dept. of ECE, Seoul National University \quad $^2$IPAI, Seoul National University \\ $^3$Adobe Research \\
\texttt{\{4bkang,minbeomkim,kms0805,drl123,hyukhunkoh-ai,kjung\}@snu.ac.kr}\\
\texttt{syoon@adobe.com}
}

\begin{document}
\maketitle
\begin{abstract}
Large Vision-Language Models (LVLMs) have demonstrated outstanding performance across various multimodal tasks. 
However, they suffer from a problem known as \textit{language prior}, where responses are generated based solely on textual patterns while disregarding image information. 
Addressing the issue of language prior is crucial, as it can lead to undesirable biases or hallucinations when dealing with images that are out of training distribution. 
Despite its importance, current methods for accurately measuring language priors in LVLMs are poorly studied.
Although existing benchmarks based on counterfactual or out-of-distribution images can partially be used to measure language priors, they fail to disentangle language priors from other confounding factors. 
To this end, we propose a new benchmark called VLind-Bench, which is the first benchmark specifically designed to measure the language priors, or \textit{blindness}, of LVLMs.
It not only includes tests on counterfactual images to assess language priors but also involves a series of tests to evaluate more basic capabilities such as commonsense knowledge, visual perception, and commonsense biases.
For each instance in our benchmark, we ensure that all these basic tests are passed before evaluating the language priors, thereby minimizing the influence of other factors on the assessment.
The evaluation and analysis of recent LVLMs in our benchmark reveal that almost all models exhibit a significant reliance on language priors, presenting a strong challenge in the field.\footnote{Evaluation code and benchmark data are available at \href{https://github.com/klee972/vlind-bench}{https://github.com/klee972/vlind-bench}.}

\end{abstract}

\section{Introduction}
\label{introduction}

Recent Large Vision-Language Models (LVLMs) have demonstrated remarkable performance across various tasks through pre-training on massive multimodal datasets and visual instruction tuning. \citep{llava, instructblip, minigpt4, mplugowl, kosmos2}. 
However, these models tend to generate responses based solely on spurious text patterns, leaving the given image unconsidered.
We refer to this problem as \textit{language prior}, borrowing the term from the Visual Question Answering (VQA) community \citep{overcoming}.
Such language priors can lead to undesirable biases \citep{visogender} and hallucinations \citep{wang2023evaluation}.
For example, when a model is presented with an image of a red banana and a yellow apple along with the question, ``Is the banana yellow?,'' it has been observed that the model frequently responds with ``Yes,'' ignoring the image content \citep{rome}.
To develop a trustworthy LVLM, resolving the language prior issue is crucial; however, it has not been explored much nor has benchmarks that can accurately measure the issues.

\begin{figure*}[t]
    \centering
    \includegraphics[width=0.92\textwidth]{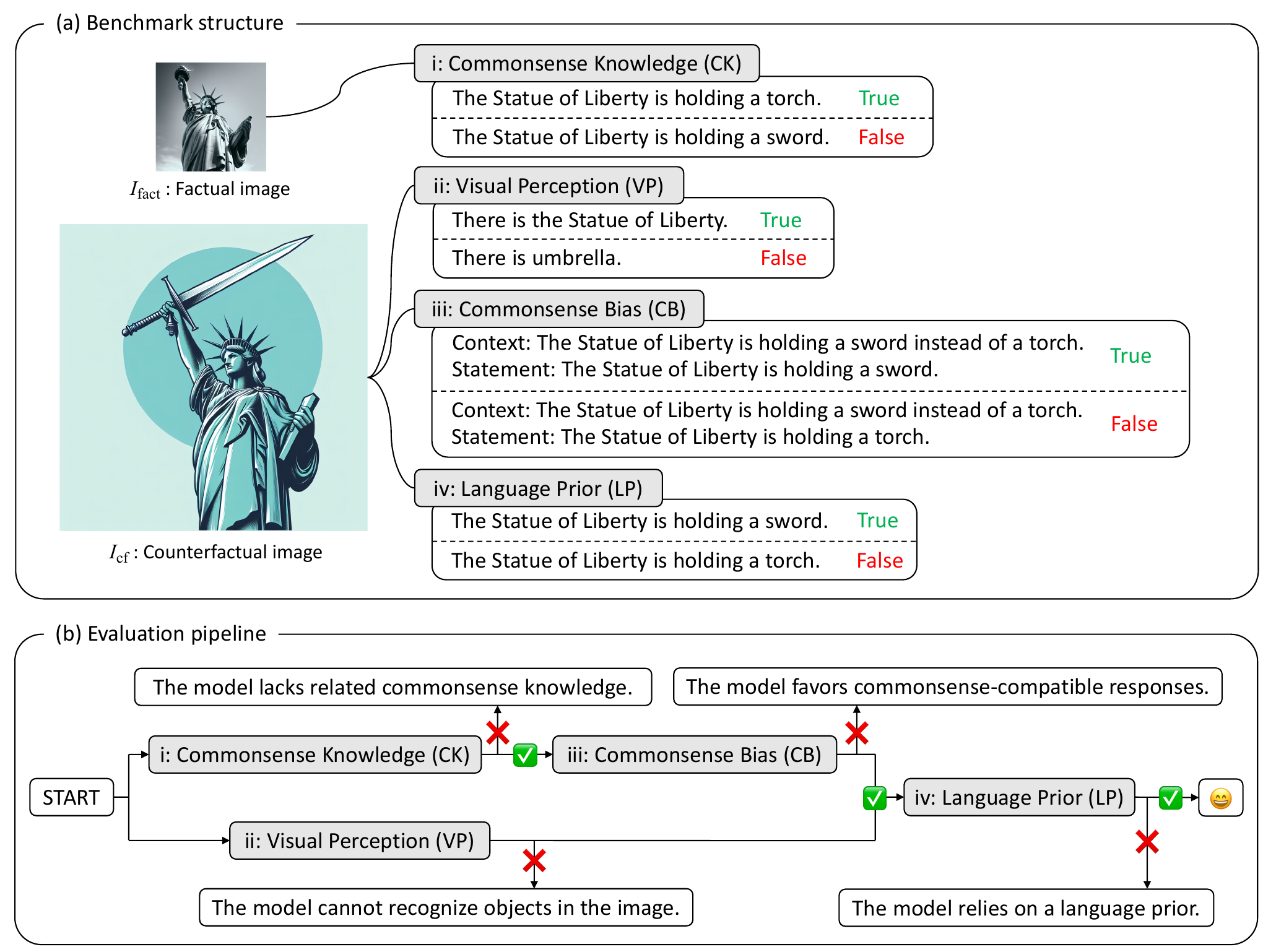}
    \caption{
    \textbf{(a)} An example from VLind-Bench. 
    Our benchmark consists of four types of questions (i-iv).
    \textbf{(b)} Evaluation pipeline of VLind-Bench.
    In the pipeline, both true and false statements of the current stage must be correctly evaluated to proceed to the next stage.
    }
    \label{figure_structure_and_pipeline}
\vspace{-8pt}
\end{figure*}

One approach to measure language priors is assessing performance on VQA benchmarks consisting of counterfactual images (e.g.,  \textsc{Whoops!} \citep{whoops} and ROME \citep{rome}).
If a model bears language priors, it will answer the question based on learned facts or common sense from its parametric knowledge without collaborating information in the given context (i.e., image); easily failing on answering counterfactual VQA tasks.
However, it is challenging to distinguish the models' misbehaviors solely caused by language priors from those caused by other deficiencies in LVLMs.
For example, there could be multiple factors affecting performance in counterfactual-contents VQA tasks -- not only language priors but also commonsense knowledge, visual perception capabilities, and the model's reluctance to counterfactual responses.
Such confounding factors make it difficult to evaluate methodologies for improving language prior problems and to assess progress in the research field.

In this paper, we propose VLind-Bench, the first benchmark that can accurately measure the language priors, or \textit{blindness}, of various LVLMs and disentangle the root causes of their failures.
To precisely measure language priors, it is necessary to create test instances that models fail \textbf{if and only if} they rely on language priors. 
For this purpose, we meticulously design a sequence of tests and measure the accuracy on each of them (Figure \ref{figure_structure_and_pipeline} (a)). 
Specifically, each instance in VLind-Bench involves four tests that can check whether a model possesses (1) commonsense knowledge, (2) visual perception, (3) commonsense bias, and (4) language prior. 
The first three serve as a sanity check performed before the test of language prior, which is the ultimate goal of our benchmark (Figure \ref{figure_structure_and_pipeline} (b)).
To the best of our knowledge, existing benchmarks can only show the individual task-level performance of LVLMs.

With VLind-Bench, we evaluate recent open-source and proprietary LVLMs' language priors.
The results show that all of the models except GPT-4o \citep{gpt4o} suffer from excessive reliance on language priors, demonstrating the challenging nature of our benchmark and the need for further improvements.
Furthermore, our experiment and analysis on existing LVLMs show that the influence of language priors is inversely proportional to the scale of the backbone LLM. 
We also reveal that Reinforcement Learning from Human Feedback (RLHF) techniques \citep{rlhfv, rlaifv}, which are designed to mitigate hallucinations, can help reduce the reliance on language priors.

\section{Related Work}
\label{related_work}
\subsection{Large Vision-Language Models}
Recently, there has been a lot of effort in extending Large Language Models (LLMs) to include visual inputs, forming a new class of models known as Large Vision-Language Models (LVLMs) \citep{llava, instructblip, minigpt4, mplugowl, gpt4v, gpt4o, gemini}.
These LVLMs are gaining attention as a new paradigm in vision-language learning by transferring the exceptional properties of LLMs, such as multi-step reasoning ability and in-context learning, to the multimodal domain.
However, these LVLMs are not free from the bias and hallucination issues inherent in LLMs \citep{visogender, li2023evaluating, Gunjal_Yin_Bas_2024, zhou2024analyzing, plausible, sgd}.
Despite this, creating benchmarks to diagnose these problems is more challenging with the image modality, leading to slower progress in benchmark development compared to LLMs.

\subsection{Benchmarks with Counterfactual Context}
Since counterfactual contexts can assess the robustness and generalization capabilities of LLMs or LVLMs, several benchmarks utilizing this approach have been proposed. 
These benchmarks assume that if a model responds based on memorized facts without properly understanding the context of text or images, it would fail to correctly solve tasks conditioned on counterfactual contexts. 
Benchmarks such as IfQA \citep{ifqa} and DisentQA \citep{disentqa} counterfactually augment textual contexts to determine whether the language model accurately incorporates augmented information when answering questions. 
\citet{reasoning-or-reciting} evaluate LLMs on reasoning tasks based on counterfactual contexts.
Benchmarks like \textsc{Whoops!} \citep{whoops}, ROME \citep{rome}, \textsc{HallusionBench} \citep{hallusionbench}, and ViLP \citep{luo2024probing} evaluate the counterfactual reasoning abilities of multimodal models by conducting VQA tasks conditioned on counterfactual images. 
However, these benchmarks cannot disentangle the reliance on language priors and commonsense biases of a model.

\section{Benchmark Structure}
\label{benchmark_structure}


VLind-Bench conducts four types of assessments, each designed to test different capabilities, as illustrated in Figure \ref{figure_structure_and_pipeline} (a). 
By providing multiple tests concerning the \textbf{exact} same image or text that are used in the language prior test, it is possible to check if the model has the essential abilities to make the language prior test meaningful. 
Depending on the problem's characteristics, each test utilizes one of two images, either factual or counterfactual, as input.

First, we provide a counterfactual image along with two statements and evaluate whether the model can correctly classify these statements as true or false based on the image (Figure \ref{figure_structure_and_pipeline} (a) - iv: Language Prior). 
If the model relies on language priors, it will not incorporate the counterfactual circumstances presented in the image into its reasoning, achieving low performance on this test.

However, merely answering questions about counterfactual images is insufficient to accurately measure the language priors due to several confounding factors. 
Firstly, when a model fails a task involving a counterfactual image, it is unclear whether this failure is due to the model's reliance on language priors or because the model possesses \textit{commonsense bias}. 
Here, commonsense bias refers to the tendency of models, including unimodal language models, to avoid responding in ways that contradict common sense. 
Therefore, we evaluate whether the model can overcome such commonsense bias \textit{regardless of modality}, by providing the model with the image and \textit{a text description of the image} as input (Figure \ref{figure_structure_and_pipeline} (a) - iii: Commonsense Bias).

Additionally, the failure in the counterfactual task might stem from an inability to recognize the objects in the counterfactual image.
Conversely, the model may simply lack common sense and pass the test merely by chance.
To this end, we provide two tests to check commonsense knowledge and visual perception abilities.
The statements used for checking commonsense knowledge are identical to those for language priors, but \textit{factual} images are given instead of counterfactual images, and the models are instructed to evaluate the truth values based on common sense (Figure \ref{figure_structure_and_pipeline} (a) - i: Commonsense Knowledge). 
In the case of visual perception, counterfactual images are still used; however, the statements are designed to assess the model's ability to recognize objects (Figure \ref{figure_structure_and_pipeline} (a) - ii: Visual Perception).

If we introspect how a human solves an counterfactual vision-language task, the three additional assessment types we propose appear more convincing. 
Humans first understand the image through visual perception, then retrieve real-world information about the objects using commonsense knowledge, and finally reason about how the given situation deviates from real-world common sense. 
Decomposing multimodal counterfactual reasoning into these three steps is a very natural approach, and each of these steps directly corresponds to our Visual Perception, Commonsense Knowledge, and Commonsense Bias tasks.

If a model fails any test assessing its basic ability, evaluating it on more complex tests that rely on that basic ability would be meaningless.
Therefore, the evaluation of our benchmark proceeds sequentially, starting with easier problems that assess fundamental abilities and gradually advancing to more difficult problems that are counterfactual and multimodal in nature (Figure \ref{figure_structure_and_pipeline} (b)).
This pipelined evaluation paradigm could be more universally applied, not only for measuring language priors but also for more accurately assessing the varying capabilities of AI systems.

\subsection{Commonsense Knowledge (CK)}
First, it is essential to verify whether the model possesses commonsense knowledge about the instances of the benchmark. 
This step allows us to determine whether the model's success at counterfactual tests is genuine or due to a lack of common sense. 
Therefore, we introduce a Commonsense Knowledge test (CK) to assess the model's commonsense knowledge about the given instances.
Specifically, the CK comprises one image $I_\textrm{fact}$ and two statements $s_\textrm{fact}$ and $s_\textrm{cf}$. 
The image $I_\textrm{fact}$ depicts a factual circumstance that aligns with common sense (e.g., an image of the Statue of Liberty). 
Among the two statements, $s_\textrm{fact}$ is a factual statement that is true based on real-world common sense (e.g., ``The Statue of Liberty is holding a torch.''), while $s_\textrm{cf}$ is a counterfactual statement that is false (e.g., ``The Statue of Liberty is holding a sword.''). 
Also, we use the prompt template, $\textrm{pr}_\textrm{CK}$, to instruct the LVLM to evaluate the truth value of the input text based on common sense.

\begin{figure}[ht]
    \centering
    \includegraphics[width=\columnwidth]{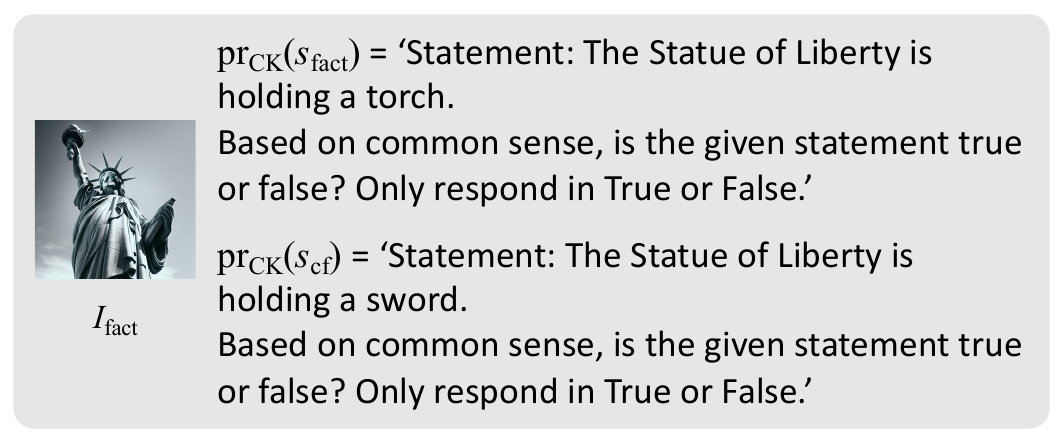}
    \label{figure_3_1}
\end{figure}
\vspace{-12pt}
To pass the CK, the model must accurately predict the truth value of both statements:
\begin{equation} \label{eq1}
    \begin{split}
        P_\textrm{CK} = &\mathbbm{1}(\textrm{LVLM}(I_\textrm{fact}, \textrm{pr}_\textrm{CK}(s_\textrm{fact}))=\textrm{``True''}\\ &\land \textrm{LVLM}(I_\textrm{fact}, \textrm{pr}_\textrm{CK}(s_\textrm{cf}))=\textrm{``False''}),
    \end{split}
\end{equation}
where $P_\textrm{CK}$ indicates whether the model passed CK or not.
$\textrm{LVLM}(i, t)$ is a composition of two functions: one that maps the image input $i$ and text input $t$ to the LVLM's response, and another that maps the LVLM's response to ``True'' or ``False'' using a string match.


\subsection{Visual Perception (VP)}
The fundamental ability underpinning all multimodal tasks is visual perception, particularly the ability to recognize objects \citep{object, monet}. 
Similar to the CK, evaluating a model on more complex tasks would be meaningless when it fails in object recognition. 
Therefore, we introduce the Visual Perception test (VP) to assess whether LVLMs can recognize objects in a given counterfactual image.
VP consists of one counterfactual image $I_\textrm{cf}$ and two statements $s_\textrm{exist}$ and $s_\textrm{nil}$.
Contrary to the CK, the image $I_\textrm{cf}$ shows a counterfactual scene, which contradicts the world knowledge or common sense (e.g., an image of the Statue of Liberty holding a sword).
The reason for using counterfactual images is that the VP needs to evaluate visual perception capabilities on the same images that are used for language prior assessments, where the use of counterfactual images is essential.

In VP, both the two statements say that ``There is \textit{object} in the image.'', while the objects are set such that $s_\textrm{exist}$ is true and $s_\textrm{nil}$ is false under the given image (e.g., ``There is the Statue of Liberty.'' and ``There is umbrella.'').
To this end, we define $P_\textrm{VP}$ to indicate whether the model passed VP, with a prompt template 
$\textrm{pr}_\textrm{VP}$ to instruct the models to evaluate the truth value of input text based on the given image.

\begin{figure}[ht]
    \centering
    \includegraphics[width=\columnwidth]{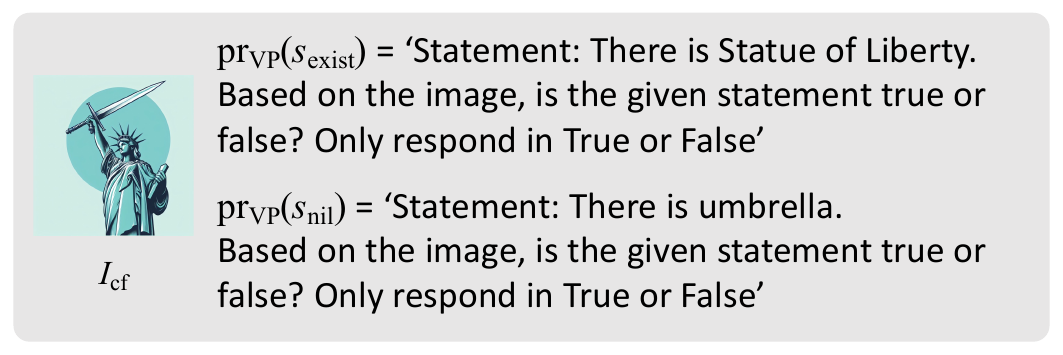}
    \label{figure_3_2}
\end{figure}
\vspace{-12pt}
The indicator for passing the VP, $P_\textrm{VP}$, is defined similarly:
\begin{equation} \label{eq2}
    \begin{split}
        P_\textrm{VP} = &\mathbbm{1}(\textrm{LVLM}(I_\textrm{cf}, \textrm{pr}_\textrm{VP}(s_\textrm{exist}))=\textrm{``True''} \\ &\land \textrm{LVLM}(I_\textrm{cf}, \textrm{pr}_\textrm{VP}(s_\textrm{nil}))=\textrm{``False''})
    \end{split}
\end{equation}

\subsection{Commonsense Bias (CB)}
It has been observed that LVLMs, including LLMs, exhibit a reluctance to provide responses that contradict common sense or learned world knowledge, even when they are explicitly instructed to respond based on counterfactual contexts \citep{whoops, rome, disentqa, ifqa}. 
We propose a Commonsense Bias test (CB) to disentangle this bias from language priors, which is the goal of this benchmark. 
To eliminate the influence of modality in the evaluation of commonsense bias, we provide LVLMs with a counterfactual textual context $T_\textrm{cf}$ and a counterfactual image $I_\textrm{cf}$ as input.
Also, we provide the models with two statements, $s_\textrm{cf}$ and $s_\textrm{fact}$, which are true and false respectively under the given context.
We wrap the context and statement with a prompt template $\textrm{pr}_\textrm{CB}$, which instructs the model to explicitly follow the information provided in the context, rather than common sense.

\begin{figure}[ht]
    \centering
    \includegraphics[width=\columnwidth]{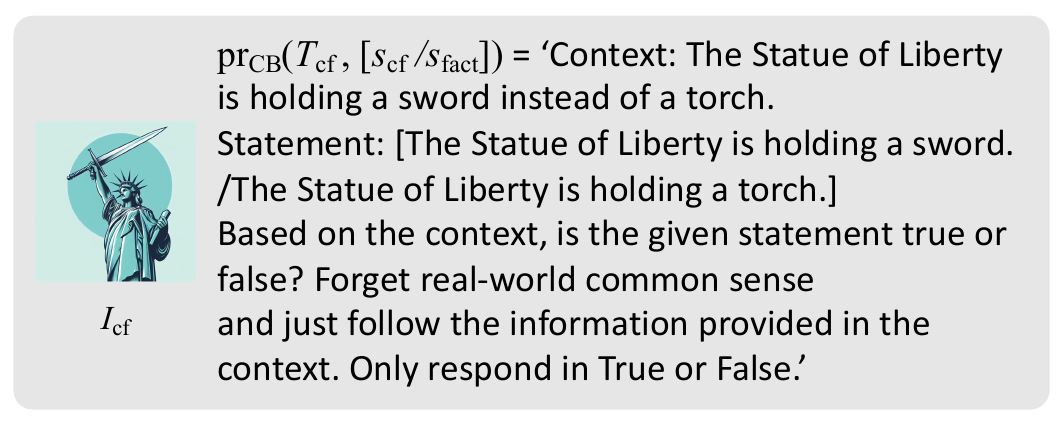}
    \label{figure_3_3}
\end{figure}
The indicator for CB is as follows:
\begin{equation} \label{eq3}
    \begin{split}
        P_\textrm{CB} = &\mathbbm{1}(\textrm{LVLM}(I_\textrm{cf}, \textrm{pr}_\textrm{CB}(T_\textrm{cf}, s_\textrm{cf}))=\textrm{``True''} \\ &\land \textrm{LVLM}(I_\textrm{cf}, \textrm{pr}_\textrm{CB}(T_\textrm{cf}, s_\textrm{fact}))=\textrm{``False''} \\ &\land P_\textrm{CK}=1)
    \end{split}
\end{equation}
Note that $P_\textrm{CB}=1$ only if $P_\textrm{CK}=1$, according to the proposed evaluation pipeline (Figure \ref{figure_structure_and_pipeline} (b)).

\subsection{Language Prior (LP)}
The evaluation of the language prior, which is the final and most crucial issue, is conducted through the Language Prior test (LP) involving a counterfactual image $I_\textrm{cf}$ and two statements $s_\textrm{cf}$ and $s_\textrm{fact}$. 
Basically, the LP is nearly identical to the CB in all aspects except for the absence of text context $T_\textrm{cf}$ and a slight difference in prompt template $\textrm{pr}_\textrm{LP}$.

\begin{figure}[ht]
    \centering
    \includegraphics[width=\columnwidth]{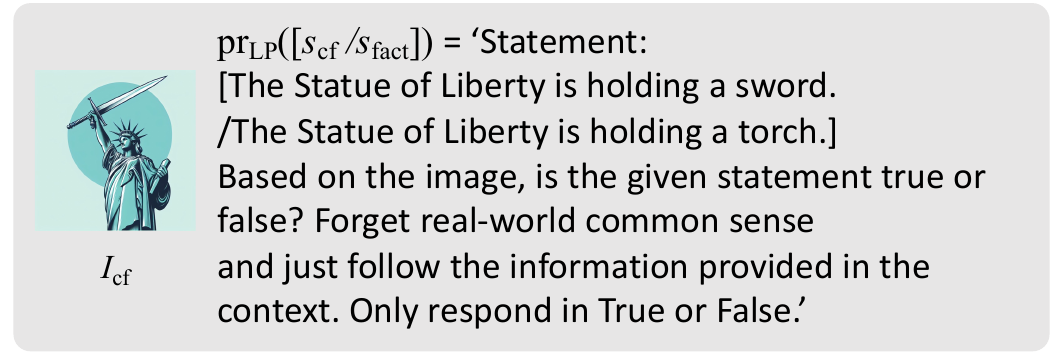}
    \label{figure_3_4}
\end{figure}
\vspace{-12pt}
The indicator for LP is as follows:
\begin{equation} \label{eq4}
    \begin{split}
        P_\textrm{LP} = &\mathbbm{1}(\textrm{LVLM}(I_\textrm{cf}, \textrm{pr}_\textrm{LP}(s_\textrm{cf}))=\textrm{``True''} \\ &\land  \textrm{LVLM}(I_\textrm{cf}, \textrm{pr}_\textrm{LP}(s_\textrm{fact}))=\textrm{``False''} \\ &\land P_\textrm{CB}=1\land\ P_\textrm{VP}=1)
    \end{split}
\end{equation}

\section{Data Generation}
\label{data_generation}
Here, we explain the data generation process of VLind-Bench.
As described in the previous section, the benchmark consists of four types of tests, incorporating various forms of images and texts. 
First, at the core of the benchmark data, there are counterfactual textual context $T_\textrm{cf}$ and image $I_\textrm{cf}$, accompanied by two statements $s_\textrm{cf}$ and $s_\textrm{fact}$, for CB and LP. 
To evaluate CK and VP, there are also a factual image $I_\textrm{fact}$ and two statements $s_\textrm{exist}$ and $s_\textrm{nil}$ regarding object recognition. 
To ensure the high quality of the data samples, we proceed with the following procedure.

\paragraph{Counterfactual Textual Contexts and Statements}

First, we generate counterfactual textual context $T_\textrm{cf}$ and corresponding statements $s_\textrm{cf}$ and $s_\textrm{fact}$, which are true and false, respectively, based on the context. 
The contexts must describe a wide range of real-world topics and be suitable for visual depiction. 
To achieve this goal, we selected 11 concepts that span various aspects of commonsense knowledge, ranging from natural sciences such as \texttt{climate} and \texttt{habitat}, to humanities such as \texttt{history} and \texttt{landmark}. 

For each selected concept, we employed GPT-4 (\texttt{gpt-4-0125-preview}) \citep{gpt4} to create 50 \textit{instance triples}, each consisting of a context, a true statement, and a false statement. 
We provided a detailed instruction with 3-shot prompt as input, using hand-crafted concept-specific examples to reflect the characteristics of each concept.

To ensure the quality of the generated data, three graduate students manually checked the correctness of the triples. We then conducted a majority vote among the three annotations to determine whether each triple should remain in our benchmark.
As a result, the initial set of 550 instance triples was reduced to 421.

\begin{table*}[t]
    \centering
    \small
    \scalebox{0.92}{
        \begin{tabular}{@{}lcccccccccccc@{}}
        \toprule
             & Climate & Color & Diet & Folklore & Habitat & History & Landmark & Location & Size & Time & Weight & Total \\ \midrule
        Num. triples &      21 &    13 &   43 &       13 &      42 &      23 &       26 &       17 &   29 &   39 &     36 &   302 \\
        Num. images  &     200 &    77 &  502 &      109 &     493 &     168 &      200 &      121 &  222 &  335 &    149 &  2576 \\ \bottomrule
        \end{tabular}
    }
\caption{The number of instance triples and images for each concept.}
\label{table_stat}
\end{table*}

\begin{table*}[t]
\centering
\small
\scalebox{0.92}{
    \begin{tabular}{@{}lccc@{}}
    \toprule
    Dataset     & Num. category/tags & Num. images & Num. image-question pairs \\ \midrule
    WHOOPS! \cite{whoops}     & 26                & 500         & 10,874                           \\
    ROME \cite{rome}        & 5                 & 1,563       & 10,941                           \\
    IfQA  \cite{ifqa}       & 7                 & -           & 6,606                            \\
    VLind-Bench & 11                & 2,576       & 14,248                           \\ \bottomrule
    \end{tabular}
}
\caption{Dataset size comparison with similar counterfactual benchmarks.}
\label{table_size_comp}
\end{table*}

\paragraph{Counterfactual Images}
Next, we proceed with the generation of counterfactual image $I_\textrm{cf}$ from the filtered textual contexts. 
Given the significance of LP in our benchmark, we generate multiple images per test for LP, unlike factual images where we generate only one image per test.
The images are generated using DALL-E 3 \citep{dalle3}, where the textual context $T_\textrm{cf}$ is provided as input, and 12 images are sampled. 
To provide diversity of image style, we produce four images each in photorealistic, illustration, and cartoon styles per one textual context. 
Consequently, for the 421 contexts, a total of 5,052 images are generated.

The generated images must provide sufficient context to accurately classify the statements as true or false and be free of artifacts. 
Similar to the previous stage, each image is verified by three graduate student reviewers and filtered using a majority vote. 
Contexts with no accepted images are also filtered at this stage. 
After this filtering process, 302 contexts and 2,274 images remained in the benchmark dataset.

\paragraph{Commonsense Knowledge and Visual Perception Tests}
In the final stage of data generation, we produce factual images $I_\textrm{fact}$ for CKs and statements $s_\textrm{exist}$ and $s_\textrm{nil}$ for VPs. 
For the factual image, since it needs to describe a circumstance where $s_\textrm{fact}$ as true, we input $s_\textrm{fact}$ directly into DALL-E 3 to generate the image. 
However, some $s_\textrm{fact}$'s are very difficult to translate into images using this method. 
In such cases, we convert $T_\textrm{cf}$ into factual textual context using GPT-4, or alternatively, we use existing images from the web.

Statements for visual perception tests are simply sentences about the presence of objects and thus can be generated using a template. 
We first prompt GPT-4 to extract one key noun from $T_\textrm{cf}$ and generate one arbitrary noun not present in $T_\textrm{cf}$. 
Then, we construct $s_\textrm{exist}$ and $s_\textrm{nil}$ using the template ``There is [\textit{noun}] in this image.''.

To verify the quality of the generated $I_\textrm{fact}$, $s_\textrm{exist}$, and $s_\textrm{nil}$, we evaluate whether OpenAI GPT-4o (\texttt{gpt-4o-2024-05-13}) \citep{gpt4o}, which is the most advanced available LVLM, passes the CK and VP. 
For instances where GPT-4o fails, human verification was conducted. 
If the failure was due to an error in the data generation process, we addressed the cause of the error by either regenerating the factual image or manually correcting the nouns in statements.

Details for human verification and input prompts are provided in Appendix \ref{appendix:HumanVerification}.

\paragraph{Statistics}
The statistics of the benchmark data generated through the process are presented in Table \ref{table_stat}. 
The difficulty of data generation varies for each concept, resulting in different proportions of samples being filtered out during the human review process. 
Ultimately, a total of 302 instance triples and 2,576 images, encompassing both counterfactual and factual images, were included in the benchmark.
We compare the size of VLind-Bench with other counterfactual benchmarks in Table \ref{table_size_comp}.
Data samples for each concept can be found in Appendix \ref{appendix:DataSamples}.

\section{Experiments}
\label{experiments}

\begin{table*}[t]
\centering
\small
\begin{tabular}{@{}lllllll@{}}
\toprule
                                & \multicolumn{4}{c}{\textbf{Pipeline Score}}                                & \multicolumn{2}{c}{\textbf{Accuracy}}    \\ \cmidrule(l){2-5} \cmidrule(l){6-7}
\textbf{Models}                                & $S_\textrm{CK}$      & $S_\textrm{VP}$      & $S_\textrm{CB}$      & $S_\textrm{LP}$      & CB          & LP          \\ \midrule
\multicolumn{7}{l}{\textit{\textbf{Proprietary LVLMs}}}                                                                                        \\
GPT-4o \citep{gpt4o}                                             & \textbf{93.0} &\textbf{96.0} &\textbf{96.8} &\textbf{89.8} &\textbf{97.0} &\textbf{89.4}          \\
GPT-4V \citep{gpt4v}                                             & 90.1 &85.4 &90.8 &77.6 &91.1 &75.6          \\
Gemini Pro Vision \citep{gemini}                                 & 80.5 &90.4 &77.0 &79.0 &75.5 &65.5          \\ \midrule
\multicolumn{7}{l}{\textit{\textbf{Open-source LVLMs}}}                                                                                       \\
LLaVA-NEXT 72B (Qwen 1.5 72B Chat) \citep{llavanext-strong}      & \textbf{94.4} &95.7 &76.1 &58.6 &75.5 &46.7          \\
LLaVA-NEXT 34B (Nous Hermes 2 Yi 34B) \citep{llavanext-strong}   & 80.5 &85.8 &61.7 &61.1 &67.2 &44.5          \\
LLaVA-1.5 13B (Vicuna v1.5 13B) \citep{liu2024improved}          & 59.9 &92.1 &40.9 &42.0 &31.5 &20.9          \\
LLaVA-1.5 7B (Vicuna v1.5 7B) \citep{liu2024improved}            & 0.0  &0.0  &-    &-    &0.0  &0.0           \\
 + RLAIF-V \citep{rlaifv}                                        & 17.9 &8.3  &48.1 &25.0 &54.3 &35.7          \\
InstructBLIP 13B \citep{instructblip}                            & 66.6 &79.5 &54.2 &57.8 &46.7 &28.0          \\
InstructBLIP 7B \citep{instructblip}                             & 58.6 &73.5 &28.2 &14.6 &27.2 &21.0          \\
OmniLMM 12B (Zephyr 7B $\beta$) \citep{rlhfv}                    & 88.1 &97.7 &\textbf{78.6} &\textbf{81.4} &\textbf{79.5} &\textbf{66.4} \\
MiniCPM-V-2 2.8B \citep{rlhfv}                                   & 76.2 &\textbf{98.3} &56.5 &68.1 &49.0 &34.1          \\
\midrule
\multicolumn{7}{l}{\textit{\textbf{Backbone LLMs}}}                                                                                                     \\
Qwen 1.5 72B Chat \citep{qwen}                                   & 75.8 & -              & 69.9          & -     & 74.2 & -              \\
Nous Hermes 2 Yi 34B \citep{nous}                                & \textbf{83.1} & -              & 75.3          & -     & \textbf{77.8} & -              \\
Vicuna v1.5 13B \citep{vicuna}                                   & 57.9& -& \textbf{80.0} & -              & 69.2          & -              \\
Vicuna v1.5 7B \citep{vicuna}                                    & 0.0          & -              & -              & -              & 0.0          & -              \\
Zephyr 7B $\beta$ \citep{zephyr}                                 & 62.3          & -              & 45.7          & -              & 40.7          & -              \\ \bottomrule
\end{tabular}
\caption{Main experimental results on VLind-Bench.}
\label{table_main}
\vspace{-6pt}
\end{table*}

\subsection{Metrics}
In section \ref{benchmark_structure}, all indicator values for the four tests have been defined for a single instance. 
For some test $\mathcal{T}\in\{\textrm{CK}, \textrm{VP}, \textrm{CB}, \textrm{LP}\}$, the final VLind-Bench score $S_\mathcal{T}$, is represented as the average of the indicator values $P_\mathcal{T}^i$'s across all instances that have passed previous tests.
\begin{equation}
    S_{\mathcal{T}}=\frac{1}{M_\mathcal{T}} \sum_{i=1}^N P_{\mathcal{T}}^i
\end{equation}
Here, $i$ is the data index, $N$ is the number of total instances in our benchmark, and $M_\mathcal{T}$ is the number of instances that have passed all the previous tests before $\mathcal{T}$ (which is essentially the number of instances considered by $\mathcal{T}$).
To be more concise, $M_\textrm{CK}=M_\textrm{VP}=N$, $M_\textrm{CB}=|\{i\mid P_\textrm{CK}^i=1\}|$ and $M_\textrm{LP}=|\{i\mid P_\textrm{CB}^i=1 \land P_\textrm{VP}^i=1\}|$.
We refer to these four scores as \textit{pipeline scores}, as they reflect the pipelined evaluation structure of VLind-Bench (columns under ``Pipeline Score'' in Table \ref{table_main}).
Alternatively, following the common definition of accuracy, the performance can be expressed as the ratio of correct instances to the total number of instances (columns under ``Accuracy'' in Table \ref{table_main}).

\subsection{Models}
We have selected and evaluated recent proprietary and open-source LVLMs on the VLind-Bench. 
The open-source LVLMs were chosen to represent a diverse range of scales and training methodologies. 
Unfortunately, the performance of the InstructBLIP models could not be evaluated using the prompt template from section \ref{benchmark_structure}, as they completely failed to generate responses. 
Therefore, we utilized a modified prompt, in which the question sentence was placed at the end. 
Additionally, we assessed the performance of some backbone LLMs on CK and CB tasks without the image input. 
To ensure the reproducibility of the experiments, all inferences were conducted under a zero temperature setting.
All the experiments are conducted using 4 NVIDIA RTX A6000 GPUs.

\subsection{Main Results}
The overall model performance is shown in Table \ref{table_main}. 
Surprisingly, numerous models demonstrated somewhat low scores in $S_\textrm{CK}$, implying a deficiency of commonsense knowledge in LVLMs. 
Conversely, $S_\textrm{VP}$ scores concerning object recognition ability exhibited relatively high scores. 
This pattern of low commonsense knowledge scores and high visual perception scores aligns with observations from previous work \citep{rome}. 
Additionally, the lower $S_\textrm{CB}$ and CB scores compared to $S_\textrm{CK}$ indicate that LVLMs are reluctant to respond contrary to commonsense knowledge. 

When comparing LP and $S_\textrm{LP}$ scores, it is evident that some models with similar LP scores exhibit differing $S_\textrm{LP}$ scores. 
For instance, while the LLaVA 1.5 13B model and the InstructBLIP 7B model have similar LP scores, the LLaVA model achieves nearly three times higher $S_\textrm{LP}$ score. 
This clear lack of correlation between LP and $S_\textrm{LP}$ scores indicates that our pipelined evaluation provides additional information beyond what can be obtained by conducting task-level evaluation alone.

Finally, the generally low $S_\textrm{LP}$ score suggests that all models, except for GPT-4o, exhibit a reliance on language priors. 
This reliance was more pronounced in open-source models compared to proprietary ones. 
The reliance on language priors appeared inversely proportional to the scale of the backbone LLM. 
This trend can be observed by comparing the $S_\textrm{LP}$ scores across various sizes of models within the same LLaVA and InstructBLIP series.

To verify the validity of the VLind-Bench, we conducted experiments on a small handcrafted evaluation set, and the results are provided in Appendix \ref{appendix:ModelPerformanceHandcrafted}.

\begin{table*}[t]
\small
\centering
\scalebox{0.85}{
    \begin{tabular}{@{}lllllllllllll@{}}
    \toprule
    Model (Score Type) & Climate & Color & Diet & Folklore & Habitat & History & Landmark & Location & Size & Time  & Weight & Total \\  \midrule
    GPT-4o ($S_\textrm{CK}$)     & 95.2    & 76.9  & 97.7 & 61.5     & 92.9    & 100.0   & 84.6     & 88.2     & 93.1 & 100.0 & 100.0  & 93.0  \\
    GPT-4o ($S_\textrm{LP}$)     & 83.3    & 93.3  & 97.1 & 91.2     & 98.2    & 92.0    & 69.7     & 100.0    & 99.2 & 100.0 & 61.0   & 89.8  \\ \midrule
    OmniLMM ($S_\textrm{CK}$)    & 100.0   & 84.6  & 97.7 & 76.9     & 92.9    & 87.0    & 92.3     & 82.4     & 41.4 & 100.0 & 94.4   & 88.1  \\
    OmniLMM ($S_\textrm{LP}$)    & 73.7    & 81.9  & 99.0 & 87.8     & 86.7    & 88.2    & 47.9     & 98.2     & 45.5 & 80.7  & 0.0    & 81.4 \\ \bottomrule
    \end{tabular}
}
\caption{Performance of selected models for different concepts.}
\label{table_concept}
\end{table*}

\paragraph{RLHF-V}
An exception to such trend between model scale and language prior is the superior performance of models that applied the RLHF-V \citep{rlhfv} methodologies. 
Models such as OmniLMM and MiniCPM trained using RLHF-V, demonstrated superior performance compared to models of similar or greater scale. 
Specifically, RLHF-V employs a method called Dense Direct Preference Optimization (DDPO) to mitigate multimodal hallucination. 
DDPO constructs win-lose pairs by having humans modify only the hallucinatory spans in the model responses to align with image information, thereby forcing the use of visual modality to increase the reward. 
Such construction of training data might be the reason for the reduced reliance on language prior. 
Additionally, the high performance of these methods on counterfactual images suggests that the ability to utilize image information generalizes to out-of-distribution samples.
Applying RLAIF-V \citep{rlaifv}, an AI-feedback variant of RLHF-V, to LLaVA 1.5 7B also results in significant performance improvement.


\paragraph{LLM performance}
Some might question whether the performance of LVLM is significantly influenced by the performance of its backbone LLM. 
To answer this question, we conducted an evaluation of several backbone LLMs on CK and CB tasks. 
The results, as illustrated in columns $S_\textrm{CK}$ and $S_\textrm{CB}$, indicate that the performance of the LLMs is not highly correlated to the performance of the LVLMs. 
Consequently, we can conclude that the absolute scale of the backbone LLMs and the training methodology have a more substantial impact on the final performance of LVLMs than the performance of the backbone LLMs themselves.

Another finding is that the LVLMs are sometimes superior to their original backbone LLMs on $S_\textrm{CB}$. 
Given that $S_\textrm{CB}$ encompasses the same content in both image and text formats, this suggests that, in certain scenarios, learning from the visual modality may enhance robustness in the text modality.

\section{Discussion}

\paragraph{Performance by Concept}
One particularly interesting finding is that the model performance varies significantly depending on the concept. 
For instance, high-performing open-source models such as OmniLMM scored zero in $S_\textrm{LP}$ for the concept of ``weight,'' and even GPT-4o only managed to achieve a score of 61.0\% (Table \ref{table_concept}). 
This suggests that although LVLMs might possess real-world knowledge about physical properties like weight, they lack robust concepts of these properties that can be generalized under counterfactual situations.

\paragraph{Chain-of-Thought Prompting}
LLMs are known to respond more comprehensively by generating intermediate reasoning steps.
In this section, we assess the effect of Zero-shot-CoT \citep{zcot} on VLind-Bench tasks by replacing the instruction in our prompts ``Only respond in True or False.'' to ``Let's think step by step.''.

\begin{table}[H]
\centering
\small
\scalebox{0.9}{
\begin{tabular}{@{}lllllll@{}}
\toprule
                                & \multicolumn{4}{c}{\textbf{Pipeline Score}}                                & \multicolumn{2}{c}{\textbf{Accuracy}}    \\ \cmidrule(l){2-5} \cmidrule(l){6-7}
\textbf{Models}                                & $S_\textrm{CK}$      & $S_\textrm{VP}$      & $S_\textrm{CB}$      & $S_\textrm{LP}$      & CB          & LP          \\ \midrule
\multicolumn{7}{l}{\textit{\textbf{Zero-shot-CoT}}}                                                                                        \\
GPT-4o                         & \textcolor{red}{91.4} & \textcolor{red}{94.7} & \textcolor{red}{93.1} & \textcolor{red}{89.4} & \textcolor{red}{93.0} & \textcolor{red}{87.8}          \\
GPT-4V                         & \textcolor{blue}{91.4} & \textcolor{blue}{95.7} & \textcolor{blue}{93.1} & \textcolor{blue}{87.1} & \textcolor{blue}{92.7} & \textcolor{blue}{85.0}    \\
LLaVA-NEXT 72B                 & \textcolor{red}{94.0} & \textcolor{red}{92.1} & \textcolor{red}{70.1} & \textcolor{blue}{72.8} & \textcolor{red}{70.9} & \textcolor{blue}{54.9}        \\
OmniLMM 12B                    & \textcolor{red}{81.5} & \textcolor{red}{94.4} & \textcolor{red}{60.2} & \textcolor{red}{50.8} & \textcolor{red}{63.2} & \textcolor{red}{35.5}          \\
MiniCPM-V-2 2.8B               & \textcolor{blue}{82.1} & \textcolor{red}{87.4} & \textcolor{blue}{63.3} & \textcolor{red}{56.4} & \textcolor{blue}{61.9} & \textcolor{blue}{37.8}      \\
\bottomrule
\end{tabular}
}
\caption{Zero-shot-CoT performance of selected models. Compared to True/False prompting, improvements are shown in \textcolor{blue}{blue}, while declines are shown in \textcolor{red}{red}.}
\label{table_cot}
\end{table}

As shown in Table \ref{table_cot}, the impact of CoT varies depending on the model type and the scores measured. 
CoT produces significant performance improvements in certain large models, particularly in $S_\textrm{LP}$ and LP scores. 
However, in other instances, the advantages of CoT are negligible, and in some cases, CoT even hinders performance. 
Additionally, we found that smaller models, such as OminLMM and MiniCPM-V-2, struggled to effectively follow CoT instructions; they generated final answer before the reasoning steps.
For these reasons, we adopted an evaluation setting that forces responses to be limited to either ``True'' or ``False.''

\paragraph{Language Priors and Model Scale}
The tendency for the language prior to be inversely proportional to the scale of backbone LLMs may appear counterintuitive (i.e., LLaVA in Table~\ref{table_main}). 
We have not identified the precise cause of this trend. 
One possible explanation is that larger pre-trained models are less prone to overfitting to the dataset during the visual instruction tuning process, thereby better maintaining their ability to attend to visual information.

In the experiments, we employ models with various scales of image encoders (ranging from approximately 300M to 5B), however, no clear correlation was observed between the language prior and the size of the image encoder.

\paragraph{Diagnosing LVLMs}
VLind-Bench can diagnose a model's capabilities in multiple aspects and components, providing clues on where to focus for comprehensive improvements. 
For instance, a low $S_\textrm{LP}$ score suggests that enhancements should be in the vision-language training aspect, while a low $S_\textrm{CK}$ score indicates that improvements should focus on the knowledge aspect of the backbone LLM.
In the case of the former, utilizing the RLHF-V techniques can significantly reduce the model's reliance on language priors, as observed in Section \ref{experiments}.

\section{Conclusion}
In this work, we proposed VLind-Bench, a benchmark designed to precisely measure language priors in LVLMs. 
We evaluated several LVLMs using this benchmark and analyzed the results, finding that the reliance on language priors is inversely proportional to the model scale. 
Additionally, the RLHF-V technique turned out to significantly aid in reducing such reliance. 
As demonstrated with VLind-Bench, we endorse a pipelined evaluation paradigm for the general construction of benchmarks to disentangle the specific abilities intended for measurement.

\section*{Limitations}
\label{limitation}
Although VLind-Bench minimized potential confounding factors in assessing language priors, there may still be unconsidered factors that contribute to the benchmark scores. 
VLind-Bench used only a single fixed prompt for evaluation, but recent studies have shown that LLMs and LVLMs respond sensitively to even small changes in such prompts \cite{para, EM}.

Additionally, the CBs in our benchmark does not necessarily need to receive both text and image as input to check the commonsense bias.
Such design choice is mostly due to a lack of established practices for feeding text-only inputs to LVLMs. 
As alternatives to $I_\textrm{cf}$, we conducted experiments using a plain single-color image or rendered text prompts as visual input (refer to Appendix \ref{appendix:PlainImage_RenderedText}); however, none of these approaches works -- these kinds of images can be considered out-of-distribution samples, and some proprietary models output error messages for these inputs.
Exploring more established methods for text-only inputs in LVLMs falls outside the scope of our paper, but further research in this area is necessary both from a practical perspective and for a deeper understanding of how individual components of LVLMs operate.

Although VLind-Bench addressed various aspects of language priors and commonsense biases, its limitation is that it did not cover social bias \cite{visogender, target} or toxicity \cite{lifetox, latte}.

Finally, we did not train the LVLMs with the data we constructed.
While our primary goal in Section \ref{data_generation} was to generate data for a benchmarking purpose, we can also use this process to generate training data automatically. 
Training LVLMs with such dataset could help mitigate reliance on language priors, but we leave this as future work.

\section*{Acknowledgments}
This work was partly supported by the National Research Foundation of Korea (NRF) grant funded by the Korea government (RS-2024-00348233), Institute of Information \& communications Technology Planning \& Evaluation (IITP) grant funded by the Korea government(MSIT) [RS-2021-II211343, Artificial Intelligence Graduate School Program (Seoul National University) \& RS-2021-II212068, Artificial Intelligence Innovation Hub (Artificial Intelligence Institute, Seoul National University)], and the BK21 FOUR program of the Education and Research Program for Future ICT Pioneers, Seoul National University in 2024.
K. Jung is with ASRI, Seoul National University, Korea.

\bibliography{custom}

\appendix

\newpage

\section{Human Verification and Model Prompt Details}
\label{appendix:HumanVerification}
\paragraph{Criteria for Instance Triple Verification}
The reviewers are provided with the context, the true statement, and the false statement (which was defined as \textit{instance triple} in the Section 4). 
For each instance triple, the reviewers are given two options: Accept and Reject.
The appropriateness is verified based on the following criteria.

\begin{mdframed}
\begin{enumerate}
    \item Decisions are made based solely on the text without considering image generation.
    \item If a true (false) statement is not clearly true (false), it should be rejected.
    \item If the context is not counterfactual, it should be rejected.
    \item Even if a true (false) statement is indeed true (false), it should be rejected if it does not address the counterfactual aspect of the context.
    \item If the truth values of statements cannot be inferred from the context, it should be rejected.
    \item Annotators may use internet searches to determine the appropriateness of the context and statement.
\end{enumerate}
\end{mdframed}

\paragraph{Criteria for Image Verification}
The reviewers are provided with the context, the true statement, the false statement, and the generated image. 
For each image, the reviewers are given two options: Accept and Reject.
The appropriateness is verified based on the following criteria.

\begin{mdframed}
\begin{enumerate}
    \item If a true (false) statement is not clearly true (false), it should be rejected.
    \item Accept the image if it is sufficient to determine the truth values of the statements, even if the image does not precisely depict the context.
    \item Reject if the generated image is of significantly poor quality.
    \item Annotators may use internet searches to determine the appropriateness of the image.
\end{enumerate}
\end{mdframed}

Each instance triple or image was reviewed by a total of three reviewers. 
Only those instance triples or images that were accepted by at least two reviewers were included in our benchmark.

\paragraph{Prompt Template for Instance Triple Generation}
We used the following prompt template for instance triple generation.
To facilitate understanding of the reader, the template is filled with examples of the concept ``location,'' with the filled-in sections indicated in \textit{italics}.

\begin{mdframed}
Given a concept, create related counterfactual situation (context) which can be described with an image. Also generate two statements with different truth values for each situation. Make only clear statements so that there is no room for vague or different truth value of the statement depending on the point of view. For example, through the concept of \textit{"location"}, we can create a counterfactual situation such as \textit{"A variety of marine life lives in the city built underwater."} and describe it with an image of \textit{a underwater city}. And then we can make two statements, \textit{"The city's buildings are surrounded by marine life."} and \textit{"The city has human residents."}, which is true and false under given counterfactual situation, respectively.
List 50 context and statement pairs for the concept of \textit{"location."} Output the results using the following json template.

\textit{[\{"id": 1, "context": "A ship is located in the middle of a large city.", "true\_statement": "The ship is surrounded by buildings.", "false\_statement": "The ship is in the ocean."\}, \{"id": 2, "context": "A glacier is found in a tropical jungle.", "true\_statement": "The glacier coexists with tropical trees.", "false\_statement": "The glacier is in the polar region."\}, ...]}
\end{mdframed}

\paragraph{Prompt Template for Generating Nouns for VPs}
As described in Section 4, we employed GPT-4 to extract one key noun from $T_{\textrm{cf}}$ and generate one arbitrary noun not present in $T_{\textrm{cf}}$, to construct $s_{\textrm{exist}}$ and $s_{\textrm{nil}}$.
To ensure appropriateness, two instances of each noun were initially generated, after which a manual selection process was conducted to choose the better option between the two.

We used the following prompt template for generating nouns for the VPs.

\begin{mdframed}
Extract nouns from the following context.
If there are more than two nouns, pick the two most important nouns.
Also generate two random nouns that are not included in the context.
Here are some examples.

Context: Wombats burrow in the frozen tundra, their tunnels creating intricate networks under the snow.
\{"nouns": ["wombat", "tunnel"], "non-existent\_nouns": ["zebra", "closet"]\}

Context: The jellybean is heavier than the digital piano.
\{"nouns": ["jellybean", "piano"], "non-existent\_nouns": ["car", "oven"]\}

Context: \textit{Context}
\end{mdframed}

\section{Experiments Using a Plain White Image and Rendered Text Prompts}
\label{appendix:PlainImage_RenderedText}

As discussed in Section 6, we conducted experiments using a plain white image and rendered text prompts as visual inputs instead of $I_{\textrm{fact}}$ and $I_{\textrm{cf}}$ in CK and CB. 
When employing the plain white image, we replaced all images in the CK and CB inputs with a plain white image. 
In the case of using rendered text prompts, we substituted all CK and CB input images with images that had the content of the textual prompts rendered in black text on a white background.

\begin{table}[H]
\centering
\small
\caption{Experimental results on VLind-Bench using various visual inputs.}
\scalebox{0.9}{
\begin{tabular}{@{}lllllll@{}}
\toprule
                                & \multicolumn{4}{c}{\textbf{Pipeline Score}}                                & \multicolumn{2}{c}{\textbf{Accuracy}}    \\ \cmidrule(l){2-5} \cmidrule(l){6-7}
\textbf{Models}                                & $S_\textrm{CK}$      & $S_\textrm{VP}$      & $S_\textrm{CB}$      & $S_\textrm{LP}$      & CB          & LP          \\ \midrule
\multicolumn{7}{l}{\textbf{$I_{\textrm{fact}}$ \textit{/} $I_{\textrm{cf}}$ \textit{as visual input}}}                                                                                        \\
GPT-4o                                             & 93.0 &96.0 &96.8 &89.8 &97.0 &89.4          \\
LLaVA-NEXT 72B                 & 94.4 &95.7 &76.1 &58.6 &75.5 &46.7          \\
OmniLMM 12B                    & 88.1 &97.7 &78.6 &81.4 &79.5 &66.4 \\ \midrule
\multicolumn{7}{l}{\textit{\textbf{plain white image as visual input}}}                                                                                        \\
GPT-4o                                             & 85.1 &96.0 &95.7 &88.4 &96.4 &89.4          \\
LLaVA-NEXT 72B                 & 88.4 &95.7 &74.9 &54.8 &74.2 &46.7          \\
OmniLMM 12B                    & 79.1 &97.7 &72.4 &81.0 &72.8 &66.4 \\ \midrule
\multicolumn{7}{l}{\textit{\textbf{rendered text prompts as visual input}}}                                                                                        \\
GPT-4o                                             & 86.1 &96.0 &96.5 &88.5 &97.0 &89.4          \\
LLaVA-NEXT 72B                 & 89.1 &95.7 &70.6 &54.2 &72.8 &46.7          \\
OmniLMM 12B                    & 74.2 &97.7 &65.2 &77.4 &68.2 &66.4 \\
\bottomrule
\end{tabular}
}
\label{table_plain}
\end{table}

Table 4 presents the results of this experiment, showing a notable performance decline, particularly in the CK. 
This performance decline can be attributed to the absence of information that was present in the original images.
Additionally, both plain white image and rendered text prompts can be considered out-of-distribution inputs (OOD), leading to unstable performance.

\section{Model Performance by Image Style}
\label{appendix:ModelPerformanceImageStyle}

Here, we observed how performance varies across different image styles. 
As mentioned in Section 4, we generated images in photorealistic, illustration, and cartoon styles.

\begin{table}[H]
\centering
\small
\caption{Experimental results on VLind-Bench with varying image styles.}
\scalebox{0.9}{
\begin{tabular}{@{}lllllll@{}}
\toprule
                                & \multicolumn{4}{c}{\textbf{Pipeline Score}}                                & \multicolumn{2}{c}{\textbf{Accuracy}}    \\ \cmidrule(l){2-5} \cmidrule(l){6-7}
\textbf{Models}                                & $S_\textrm{CK}$      & $S_\textrm{VP}$      & $S_\textrm{CB}$      & $S_\textrm{LP}$      & CB          & LP          \\ \midrule
\multicolumn{7}{l}{\textbf{\textit{photorealistic}}}                                                                                        \\
GPT-4o                                             & 93.1 & 96.2 & 97.1 & 92.3 & 97.3 & 91.6          \\
LLaVA-NEXT 72B                 & 94.6 & 95.8 & 77.2 & 65.0 & 76.5 & 52.4          \\
OmniLMM 12B                    & 88.8 & 97.7 & 81.8 & 82.8 & 83.1 & 70.5          \\ \midrule
\multicolumn{7}{l}{\textit{\textbf{illustration}}}                                                                                        \\
GPT-4o                                             & 92.7 & 95.4 & 97.5 & 90.1 & 97.7 & 90.0          \\
LLaVA-NEXT 72B                 & 94.3 & 96.2 & 78.5 & 59.1 & 77.8 & 47.3          \\
OmniLMM 12B                    & 88.5 & 98.1 & 81.4 & 80.4 & 82.4 & 67.7          \\ \midrule
\multicolumn{7}{l}{\textit{\textbf{cartoon}}}                                                                                        \\
GPT-4o                                             & 94.1 & 96.7 & 97.2 & 91.9 & 97.4 & 91.5          \\
LLaVA-NEXT 72B                 & 94.8 & 95.5 & 78.8 & 58.2 & 78.8 & 48.0          \\
OmniLMM 12B                    & 87.7 & 97.8 & 82.2 & 82.0 & 82.5 & 68.4          \\
\bottomrule
\end{tabular}
}
\label{table_style}
\end{table}

Table 5 shows that the performance across these styles in the CK, VP, and CB did not vary significantly. 
A notable variation in performance was observed only in LP, where the photorealistic style yielded better results compared to the other two styles. 
This could be due to the model's assessment that images in the illustration or cartoon styles lack realism compared to photorealistic images, leading it to generate responses that align more closely with common sense.

\section{Model Performance on Handcrafted Evaluation Set}
\label{appendix:ModelPerformanceHandcrafted}

To verify the validity of automatically generated text and images, we created a small handcrafted evaluation set and assessed the performance of several models, comparing it with their performance on the original VLind-Bench. 
All the text in the handcrafted evaluation set was written by humans, with three “triples” for each concept. 
For each triple, we generated three counterfactual images and one factual image. 
It was extremely challenging to find real images that depict counterfactual situations, and even if such images were found online, there was no way to verify whether they were outputs from generative models.
To eliminate any potential advantage that OpenAI models might have from using DALL-E 3-generated images, we generated all the images using Stable Diffusion and Adobe Firefly, incorporating various styles such as photorealistic, illustration, and cartoon. 
This handcrafted evaluation set ultimately consists of 33 triples and 132 images, and the performance on this set is as follows.

\begin{table}[H]
\centering
\small
\caption{Experimental results on handcrafted evaluation set.}
\scalebox{0.87}{
\begin{tabular}{@{}lllllll@{}}
\toprule
                                & \multicolumn{4}{c}{\textbf{Pipeline Score}}                                & \multicolumn{2}{c}{\textbf{Accuracy}}    \\ \cmidrule(l){2-5} \cmidrule(l){6-7}
\textbf{Models}                                & $S_\textrm{CK}$      & $S_\textrm{VP}$      & $S_\textrm{CB}$      & $S_\textrm{LP}$      & CB          & LP          \\ \midrule
GPT-4o                         & 90.9 & 90.9 & 100.0 & 93.8 & 100.0 & 93.9          \\
GPT-4V                         & 90.9 & 81.8 & 90.0 & 75.8 & 87.9 & 72.7          \\
LLaVA-NEXT 72B                 & 87.9 & 97.0 & 79.3 & 60.9 & 78.8 & 47.5          \\
OmniLMM 12B                    & 81.8 & 97.0 & 85.2 & 76.8 & 87.9 & 70.7          \\
MiniCPM-V-2 2.8B               & 63.6 & 97.0 & 66.7 & 50.0 & 57.6 & 39.4          \\ 
\bottomrule
\end{tabular}
}
\label{table_handcrafted}
\end{table}

As shown in Table \ref{table_handcrafted}, the performance on the original VLind-Bench and the gold evaluation set does not differ significantly (refer to Table \ref{table_main} for the original VLind-Bench scores).

\section{Data Samples}
Here, we provide some data samples for each concept (next page).
For the notations, please refer to the section \ref{benchmark_structure}.
\label{appendix:DataSamples}

\begin{figure*}[ht]
    \centering
    \includegraphics[width=0.94\textwidth]{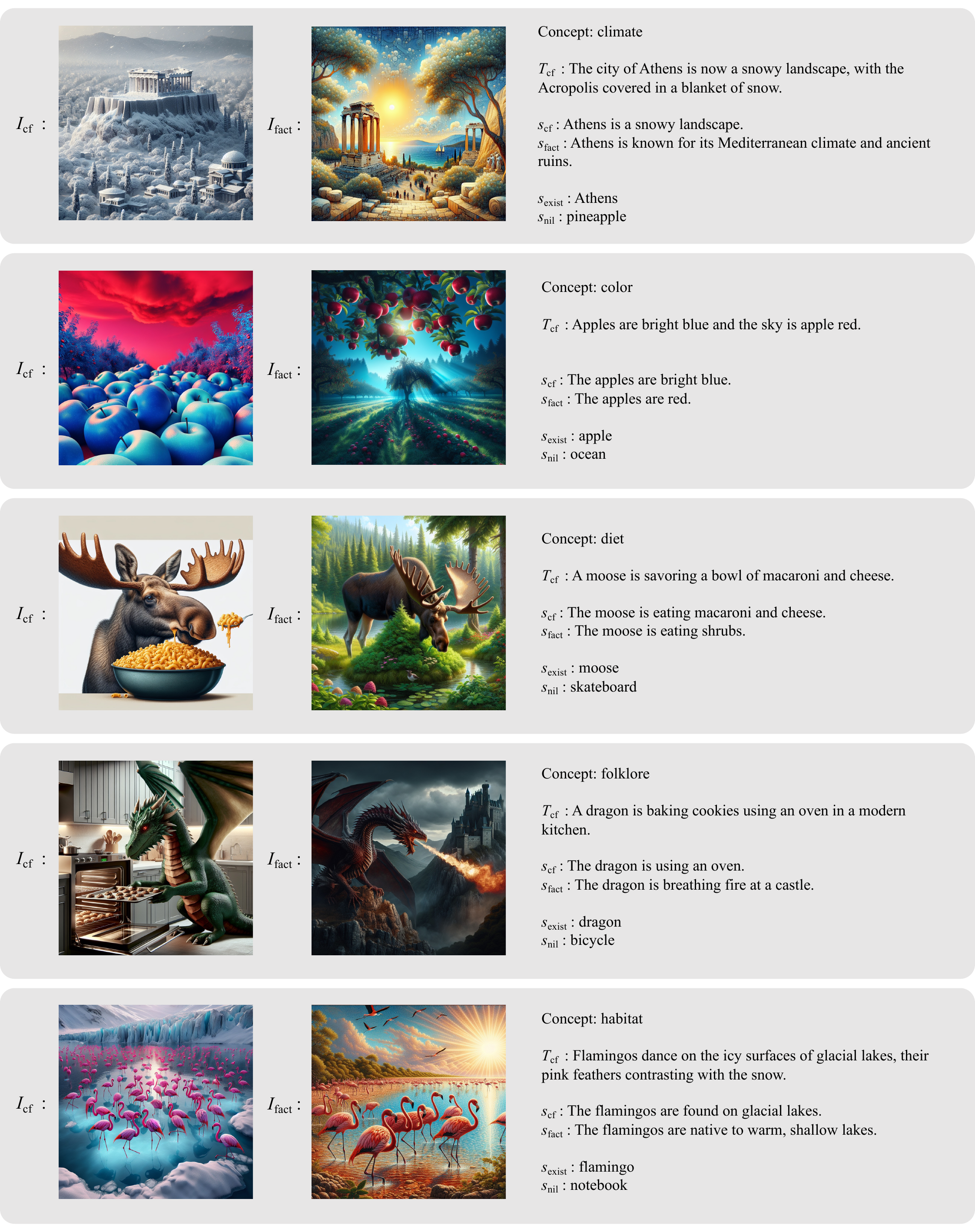}
    \caption{Data samples for concept of climate, color, diet, folklore, and habitat.}
\end{figure*}
\begin{figure*}[t!]
    \centering
    \includegraphics[width=0.94\textwidth]{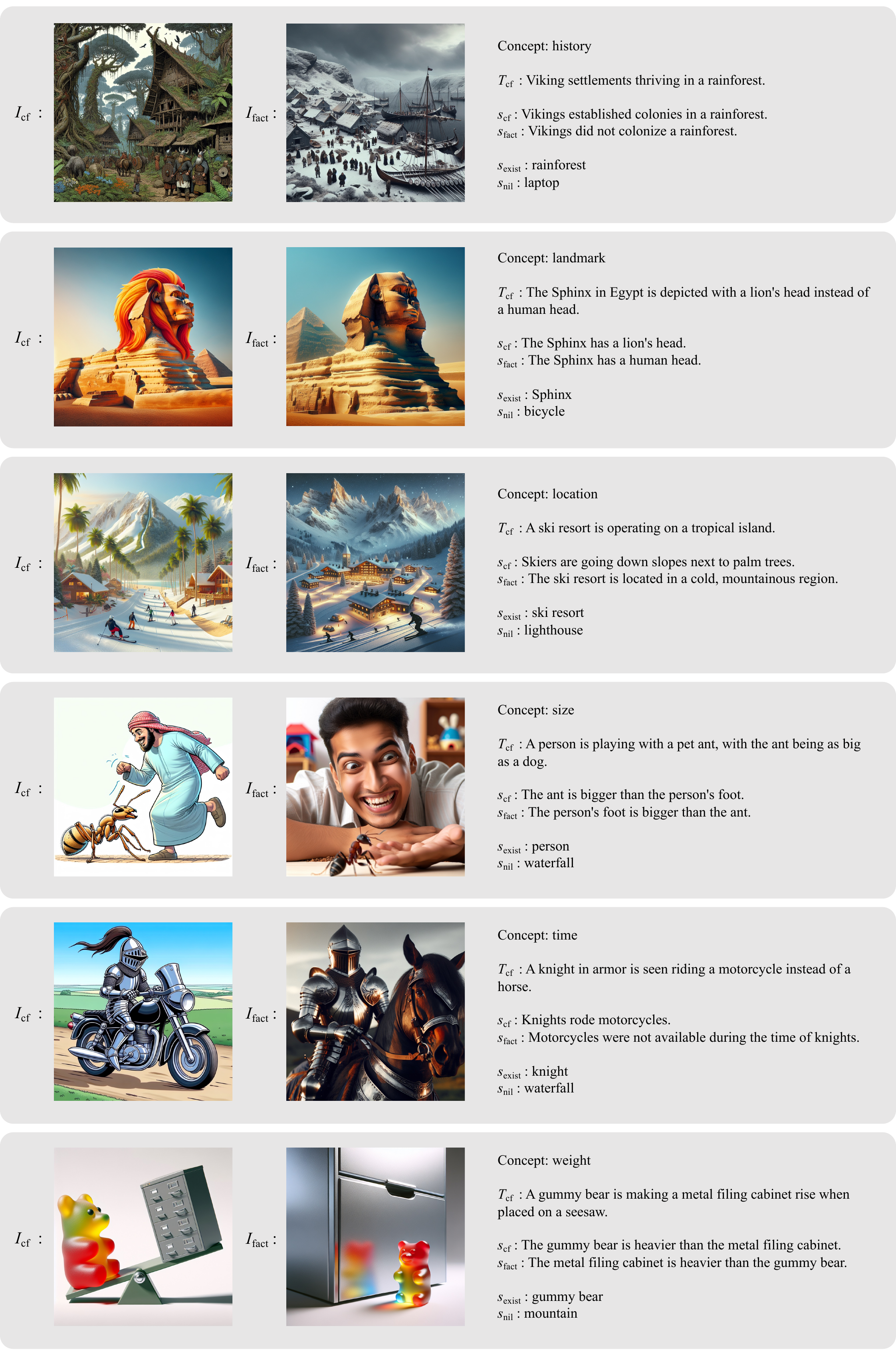}
    \caption{Data samples for concept of history, landmark, location, size, time, and weight.}
\end{figure*}

\section{Information About Use Of AI Assistants}

In writing this paper, we utilized ChatGPT \footnote{https://chatgpt.com/} for paraphrasing.

\end{document}